\documentclass{article}

\usepackage[utf8]{inputenc} 
\usepackage[T1]{fontenc}    

\usepackage{url}            
\usepackage{booktabs}       
\usepackage{amsfonts}       
\usepackage{nicefrac}       
\usepackage{microtype}      
\usepackage{xcolor}         
\usepackage[pdftex]{graphicx}
\usepackage{amsmath}
\usepackage{amssymb}
\usepackage{mathtools}
\usepackage{amsthm}
\usepackage{subcaption}
\usepackage{multirow}
\usepackage{makecell}
\usepackage{graphics, enumitem}
\usepackage{float}

\usepackage{caption}
\usepackage{xspace}
\usepackage{hyperref}

\usepackage[preprint]{icml2026}

\def\eg{{e.g.}}

\def\sim{\mathrm{sim}}

\usepackage{pifont}
\usepackage{tikz}
\newcommand*\circled[1]{\tikz[baseline=(char.base)]{
            \node[shape=circle,draw,inner sep=1pt] (char) {#1};}}

\newenvironment{denseitemize}{
\begin{itemize}[topsep=2pt, partopsep=0pt, leftmargin=1.5em]
  \setlength{\itemsep}{2pt}
  \setlength{\parskip}{0pt}
  \setlength{\parsep}{0pt}
}{\end{itemize}}

\def\name{tLoRA\xspace}

\icmltitlerunning{tLoRA: Efficient Multi-LoRA Training with Elastic Shared Super-Models}

\begin{document}

\twocolumn[
\icmltitle{\name: Efficient Multi-LoRA Training with Elastic Shared Super-Models}

	\begin{icmlauthorlist}
		\icmlauthor{Kevin Li}{uiuc}
		\icmlauthor{Dibyadeep Saha}{uiuc}
		\icmlauthor{Avni Kanodia}{uiuc}
		\icmlauthor{Fan Lai}{uiuc}
	\end{icmlauthorlist}

	\icmlaffiliation{uiuc}{Siebel School of Computing and Data Science, University of Illinois Urbana-Champaign, Champaign, Illinois, United States}

	\icmlcorrespondingauthor{Fan Lai}{fanlai@illinois.edu}


	\vskip 0.3in
]

\printAffiliationsAndNotice{}

\begin{abstract}
As Low-Rank Adaptation (LoRA) becomes the standard approach for efficiently fine-tuning large language models (LLMs), shared clusters increasingly execute many concurrent LoRA training jobs over the same frozen backbone. While recent advances enable batching (co-locating) multiple adapters during serving, efficient training-time co-location of heterogeneous LoRA adapters presents unique challenges. Jobs often differ in adapter rank, batch size, and resource allocation, and naïve batching can introduce synchronization stalls, communication overheads, and per-job slowdowns that are worse than executing independently.
We introduce \name, a framework that enables efficient batch training of multiple LoRA jobs. \name fuses adapters that share the same base model into an elastic shared super-model, exploiting existing distributed training frameworks to derive parallelism plans that share resources effectively. At the kernel level, \name employs a fused LoRA kernel that adaptively reconstructs low-rank computation tiles and schedules rank-aware nano-batches to maximize overlap between computation and communication across adapters. At the scheduling layer, \name incorporates an online, residual-capacity-aware scheduler that adaptively groups jobs to maximize collective throughput. 
Evaluations using real-world cluster traces demonstrate that \name improves training throughput by 1.2--1.8x, job training completion time by 2.3--5.4x, and GPU utilization by 37\%.

\end{abstract}

\section{Introduction}
\label{sec:intro}

Low-Rank Adaptation (LoRA) has emerged as a lightweight and effective technique for adapting large language models (LLMs) to diverse downstream tasks. Rather than fine-tuning all model parameters, LoRA freezes the pre-trained backbone and inserts small, trainable low-rank matrices into selected attention and/or projection layers, typically accounting for less than 5\% of the total model parameters~\cite{hu2021lora}. During training, only adapter parameters are updated, substantially reducing overhead.

Increasingly, LoRA fine-tuning has become a prominent workload in modern machine learning (ML) clusters. For example, platforms such as CivitAI host over 100K LoRA adapters for a wide range of models~\cite{stylus-arxiv24}. Recent studies further suggest that fine-tuning jobs constitute a substantial fraction of shared ML cluster workloads~\cite{llmtrace-nsdi24}. These LoRA tuning tasks, submitted by individual or collaborative developers and teams, may explore different hyperparameter configurations~\cite{adalora-iclr2023} or continuously adapt models to evolving data and tasks~\cite{sdlora-iclr2025}.

Supporting large numbers of LoRA adapters at scale introduces significant efficiency challenges. Treating each adapter independently requires replicating the base model across multiple GPUs, leading to considerable memory and compute overhead. Recognizing that many adapters share the same backbone (e.g., Qwen3 model), recent advances~\cite{slora-mlsys24, dlora-osdi24} have explored batching multiple LoRAs during \emph{serving}. By allowing adapters to share a single base model and execute jointly—synchronizing after each layer—these approaches reduce redundant memory usage and improve GPU utilization. However, despite their success at inference time, the problem of efficiently \emph{training} heterogeneous LoRA adapters together remains largely unexplored.

Despite the potential efficiency gains, batching (co-locating) multiple LoRA training jobs introduces several significant challenges.
(i)~\emph{Job heterogeneity}: tuning jobs differ in LoRA rank (e.g., adapter size), batch size, and allocated resources. Naively batching jobs—especially those that are already compute-saturated—can degrade collective throughput by amplifying communication overheads and exacerbating imbalance.
(ii)~\emph{Batch execution interdependence}: executing batches over a shared base model can improve hardware utilization, but it also introduces per-layer synchronization stalls.
(iii)~\emph{Per-job incentives}: although batching can increase aggregate throughput, individual jobs—particularly those with abundant resources—may experience slowdowns under shared execution. Without explicit incentive and fairness guarantees, such degradation discourages users from participating in batched training (\S\ref{sec:background}).

We introduce \name, a framework for efficiently batching heterogeneous LoRA training jobs in shared clusters to improve collective training efficiency while respecting per-job progress requirements (e.g., completion deadlines).
To achieve this, \name first fuses multiple heterogeneous LoRA jobs into an \emph{elastic Shared Super-Model (SSM)} graph, where nodes represent computation associated with base-model layers and LoRA adapters, and edges capture data-flow (e.g., activation) dependencies.
This unified representation allows \name to seamlessly leverage existing distributed frameworks (e.g., Megatron-LM~\cite{megatron-sc21} or PyTorch FSDP) to derive efficient execution plans across GPUs (e.g., pipeline parallelism).

To mitigate execution stragglers arising from LoRA heterogeneities, \name further partitions inputs from different LoRA jobs into nano-batches, enabling fine-grained overlap between computation and communication and reducing pipeline bubbles during execution.
At runtime, \name dynamically determines which jobs to batch based on their residual resource capacity (e.g., unused GPU compute or memory when running independently), selecting groupings that maximize aggregate throughput while enforcing per-job progress guarantees, such as starvation avoidance and bounded slowdown. Our evaluations using real-world cluster traces demonstrate that tLoRA improves training
throughput by 1.2--1.8x, per-job completion time by 2.3-5.4x, and GPU utilization by up to 37\% (\S\ref{sec:eval}).

Overall, we make the following contributions:
\begin{denseitemize}
    \item We introduce a new SSM abstraction to unify LoRA training jobs, enabling efficient distributed execution. 
    
    \item We design a fused LoRA kernel that dynamically allocates low-rank computation tiles and schedules rank-aware nano-batches to maximize resource utilization.
    
    \item We propose an online, residual-capacity-aware scheduling algorithm that adaptively groups LoRA jobs to improve throughput while respecting per-job progress.
    
    \item We demonstrate the effectiveness of \name through real-world deployments.
\end{denseitemize}

\section{Background and Motivation}
\label{sec:background}

\begin{figure}[t]
    \centering
    \begin{minipage}[t]{0.46\linewidth}
        \vspace{0pt}
        \centering
        \includegraphics[width=\linewidth]{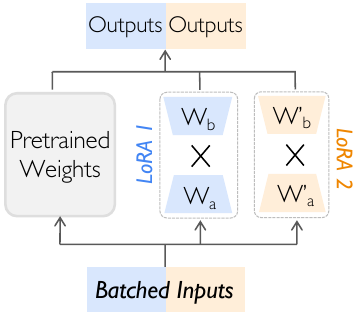}
        \caption{Adapter heterogeneity (e.g., in rank and batch size) creates tension between throughput and per-job latency in multi-LoRA training. }
        \label{fig:dlora-paradigm}
    \end{minipage}
    \hfill
    \begin{minipage}[t]{0.50\linewidth}
        \vspace{0pt}
        \centering
        \includegraphics[width=.9\linewidth]{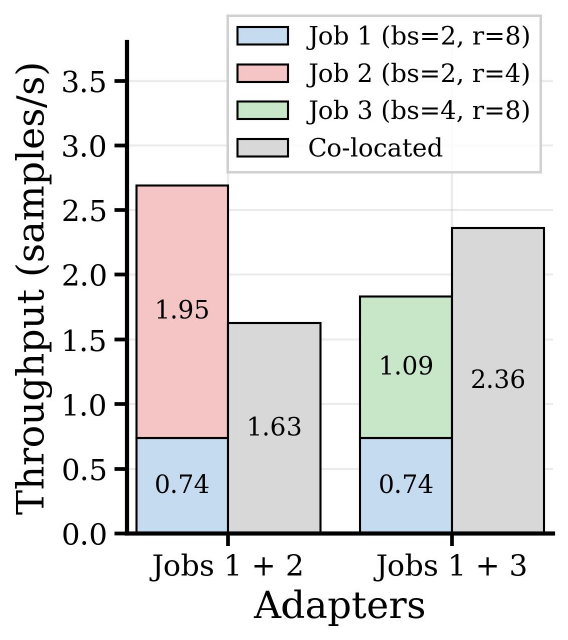}
        \vspace{-.2cm}
        \caption{Na\"ive batch LoRA training may hurt aggregate training throughput. (Llama3.1-8B)}
        \label{fig:motivating-exp}
    \end{minipage}
    \vspace{-.3cm}
\end{figure}


LoRA freezes the pre-trained backbone and inserts lightweight, trainable low-rank matrices into selected layers. As shown in Figure~\ref{fig:dlora-paradigm}, a weight update $W \leftarrow W + \Delta W$ is reparameterized as $\Delta W = A B^{\top}$, where $A \in \mathbb{R}^{d \times r}$ and $B \in \mathbb{R}^{r \times k}$ with $r \ll \min(d, k)$. This design slashes the number of trainable parameters from $\mathcal{O}(dk)$ to $\mathcal{O}(r(d + k))$. 
Because many adapters share the same frozen backbone, recent serving-time systems such as dLoRA~\cite{dlora-osdi24} and S-LoRA~\cite{slora-mlsys24} batch (co-locate) multiple LoRAs as plug-ins to a shared base model. By batching base-model computation and fusing adapter execution, grouped execution amortizes kernel-launch and I/O overheads, thereby improving inference efficiency.

Extending adapter batching from inference to training introduces unique challenges arising from along three key dimensions: (1) \emph{LoRA Heterogeneity}: varying adapter ranks, batch sizes, and sequence lengths, which shape gradient computation load; (2) \emph{Resource Heterogeneity}: different accelerator counts or even types; and (3) \emph{User-demand Heterogeneity}: divergent user's requirements on performance such as job completion time or latency constraints in online learning.  
These factors jointly determine both step efficiency and end-to-end convergence behavior.

\paragraph{Grouped Training Introduces Execution Bubbles.}
Despite its potential, naïve batching can be ineffective and even counterproductive. As illustrated in Figure~\ref{fig:motivating-exp}, some groupings of LoRA jobs improve aggregate training throughput relative to isolated execution. Merging and batching LoRA Jobs~1 and~3 increases throughput from $0.74+1.09$ (isolated runs) to $2.36$ when batched. However, other groupings lead to clear regressions (e.g., Jobs~1 and~2). This discrepancy stems from the fact that batching pools accelerators and distributes execution across devices, yet naïve strategies ignore heterogeneity in adapter rank, batch size, and device placement, potentially introducing imbalance in model parallelism, amplifying synchronization delays that shift the performance bottleneck from computation to communication, especially when jobs are grouped across nodes.

This makes job grouping a nontrivial optimization problem. Naïve grouping can cause resource-rich jobs to subsidize others and suffer slower progress, discouraging participation. With proper coordination, however, grouping enables dynamic resource reallocation: jobs with slack can temporarily yield capacity to accelerate others and later reclaim resources to reduce their own time-to-convergence.

\section{\name Design}
\label{sec:overview}

We present \name, a heterogeneity-aware framework for jointly training multiple LoRA adapters with high throughput while respecting individual job performance requirements (e.g., completion deadlines or throughput targets).

\subsection{\name Overview}
\begin{figure}[h]
    \centering
    \includegraphics[width=0.99\linewidth]{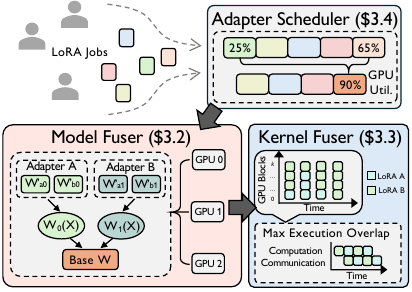}
    \caption{Lifecycle of multi-adapter LoRA training with \name.}
    \label{fig:lifecycle}
\end{figure}

\name operates in an online setting where LoRA tuning jobs continuously arrive and complete over time. At its core, \name builds on a \emph{Shared Super-Model (SSM)} abstraction that unifies a single base LLM with multiple attached adapters into a single executable model. This abstraction preserves compatibility with existing distributed training stacks (e.g., model-parallel frameworks) while enabling efficient resource sharing across heterogeneous jobs. 

Using the SSM, \name coordinates cluster-wide LoRA training through three key components.
As illustrated in Figure~\ref{fig:lifecycle}, for each group of jobs, \circled{1}~\emph{Model Fuser} (\S\ref{subsec:fuser}) consolidates the shared backbone and selected adapters into an SSM. The fused model captures structural heterogeneity across jobs (e.g., adapter rank, batch size, and sequence length) and is passed to an existing parallelism planner, which derives a distributed execution plan that minimizes pipeline bubbles and improves hardware utilization. 
During execution, \circled{2}~\emph{Kernel Fuser} (\S\ref{subsec:kernel}) performs adapter-aware kernel fusion within the SSM. It applies load balancing across streaming multiprocessors (SMs), enabling effective overlap between computation and communication and mitigating inefficiencies caused by small or irregular kernels. 
Finally, \circled{3}~\emph{Adapter Scheduler} (\S\ref{subsec:job-scheduler}) monitors lightweight runtime signals such as per-job training progress and iteration latency. At the end of each scheduling horizon, it adaptively updates grouping decisions for subsequent execution: jobs whose progress slows beyond acceptable bounds are decoupled or rebalanced, while compatible jobs are merged to improve aggregate throughput and resource efficiency.

We next describe how \name achieves high collective training throughput by minimizing execution bubbles across the model-parallel pipeline (\S\ref{subsec:fuser}), runtime kernels (\S\ref{subsec:kernel}), and adaptive training grouping decisions (\S\ref{subsec:job-scheduler}).

\subsection{Model Fuser: Optimizing Model Parallelism}
\label{subsec:fuser}

Batch training of multiple LoRA adapters offers opportunities to amortize backbone computation, I/O overhead (e.g., loading model weights to GPU register), and improve accelerator utilization, but heterogeneity in adapter rank, batch size, and device placement can introduce significant execution imbalance (\S\ref{sec:background}). 
\name addresses this challenge with a \emph{heterogeneity-aware Model Fuser} that unifies multiple training jobs into a single optimizable model representation, enabling joint reasoning about model parallelism and downstream overlap opportunities.

\paragraph{Shared Super-Model Representation}
\name introduces a \emph{Shared Super-Model (SSM)} abstraction that consolidates a set of LoRA training jobs sharing the same frozen backbone into one composite computation graph.
Given jobs $\mathcal{J}=\{J_1,\ldots,J_K\}$ derived from a base model $M$, the Model Fuser performs layer-wise architectural fusion on shared backbone operators.
LoRA adapters are retained as lightweight, job-specific branches attached to the fused backbone layers.
This design enables multiple jobs to share compute-intensive backbone execution while preserving independent forward/backward semantics and optimizer states.
The resulting SSM is functionally equivalent to training each job independently, ensuring correctness and convergence, while exposing more structure for optimization.

Rather than introducing a new model-parallel planner, \name presents the SSM as a single composite model to existing planning frameworks (e.g., PyTorch Distributed, Metis~\cite{metis-atc24}).
Through standard layer-wise profiling and cost modeling, these planners naturally internalize load heterogeneity across adapters (e.g., due to different batch sizes) when estimating per-layer compute and communication costs.
As a result, partitioning and placement decisions directly embed adapter heterogeneity into the execution plan, yielding model-parallel strategies that jointly optimize collective throughput and resource utilization without modifying existing infrastructures.

\subsection{Kernel Fuser: Minimizing Runtime Bubbles}
\label{subsec:kernel}

While the Model Fuser mitigates imbalance at the level of model parallelism, batched execution introduces additional challenges at the GPU execution level.
When heterogeneous adapters are executed jointly, differences in rank, sequence length, and batch size lead to skewed workloads across threads and warps, resulting in synchronization stalls and underutilized streaming multiprocessors (SMs). 
A naïve design that processes each adapter independently launches one kernel per adapter, resulting in many GPU kernels, incurring excessive overhead, poor occupancy, and underutilizing GPU parallelism. 
Conversely, forcing all adapters into a single dense matrix multiplication creates a ``super-kernel'' with highly irregular, block-sparse layouts with zero regions depending on the adapter heterogeneity, wasting compute and memory bandwidth.

\begin{figure}[t]
    \centering
    \includegraphics[width=.8\linewidth]{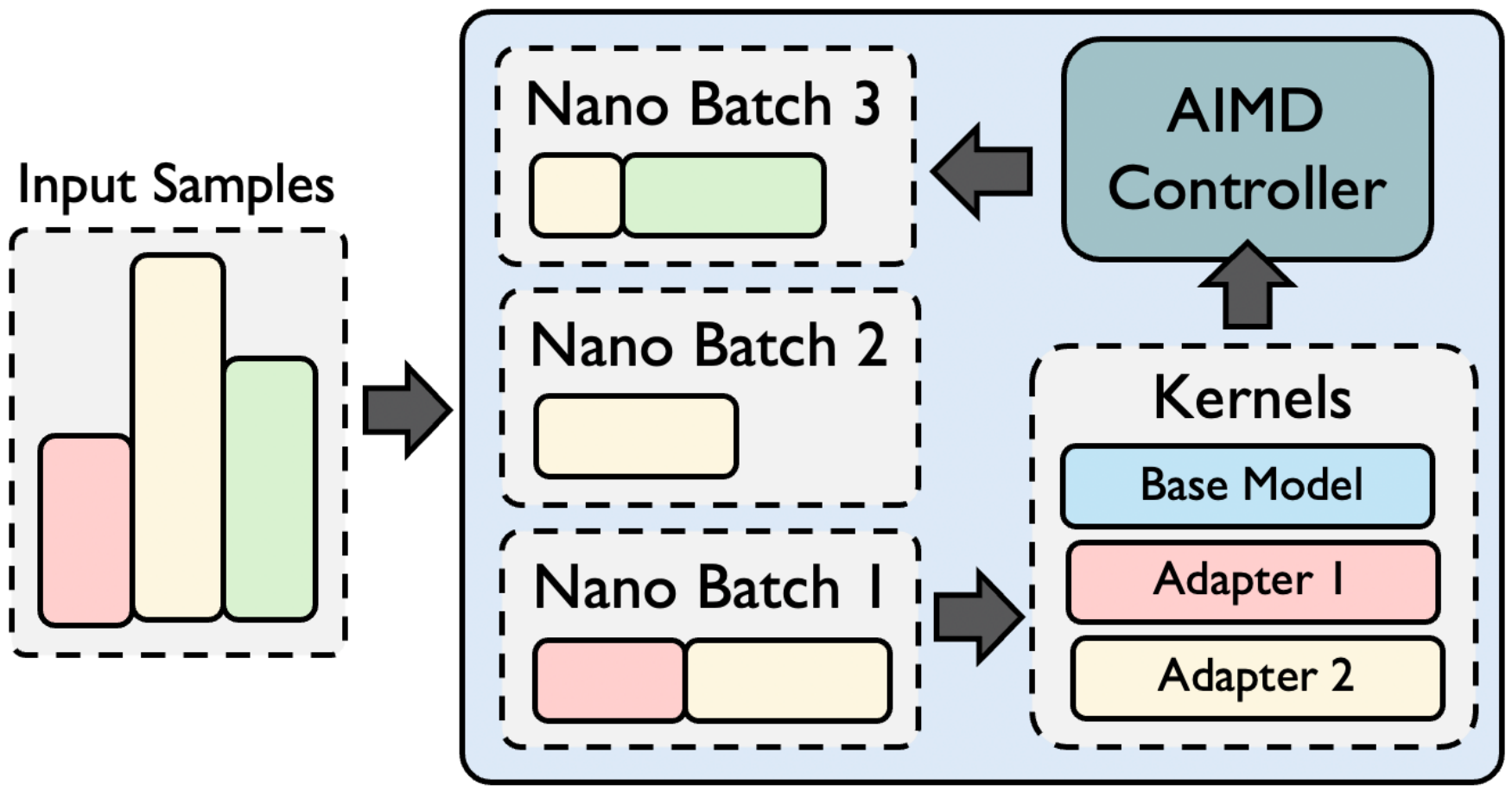}
    \caption{Execution of a micro-batch using our Kernel Fuser}
    \label{fig:kernel-fuser}
\end{figure}

As shown in Figure~\ref{fig:kernel-fuser}, to efficiently execute heterogeneous LoRA adapters, \name introduces a \emph{fused batched LoRA kernel} that avoids materializing adapter-specific weight matrices, thereby improving register and shared-memory reuse. For each adapter $i$, tokens mapped to that adapter are first gathered and multiplied with the down-projection matrix $A_i$, producing a compact intermediate of shape $(|X_i|, r_i)$. This intermediate is then immediately multiplied with the corresponding up-projection matrix $B_i$ and scattered back to the output tensor, without ever materializing $W_i = A_i B_i^\top$ or allocating full-sized temporary buffers.
Leveraging our SSM representation, \name is able to exploit Triton’s auto-tuning to select adapter-aware block sizes and tiling strategies, maximizing compute utilization across heterogeneous adapters.

\paragraph{Adaptive Nano-Batching.}
Despite improved compute balance from kernel-level fusion, batched LoRA training frequently incurs cross-GPU communication (e.g., when pooling accelerators across multiple jobs). In such settings, end-to-end efficiency is often limited not by raw compute throughput, but by how effectively computation can be overlapped with communication along the iteration critical path.

To expose fine-grained overlap opportunities, \name introduces a \emph{nano-batch} abstraction at kernel execution time. A nano-batch partitions the current training batch—either a mini-batch or a micro-batch under pipeline parallelism—along the batch dimension into smaller execution units. Concretely, for a group of LoRA jobs with batch sizes $\{B_i\}$, \name divides their combined batch into $N$ nano-batches, each containing approximately $\sum_i B_i / N$ samples. Samples in a nano-batch are processed together by the fused LoRA kernel before proceeding to the next nano-batch.

Choosing the appropriate nano-batch size is non-trivial. With a too large nano-batch size, computation proceeds in long phases that delay communication, reducing overlap; with a too small nano-batch size, kernel launch overhead dominates and compute resources become underutilized. The optimal nano-batch size for maximum compute and communication overlap may also vary depending on the accelerator architecture and the inter-GPU connection bandwidth of the cluster, which is difficult to predict analytically without runtime profiling or empirical tuning. From an optimization perspective, the iteration time can be viewed as
\begin{equation}
T_{\text{iter}} \approx \max\!\left(\sum_{n=1}^{N} T_{\text{comp}}(n),\; \sum_{n=1}^{N} T_{\text{comm}}(n)\right),
\end{equation}
where effective pipelining aims to minimize the critical path by overlapping $T_{\text{comp}}$ and $T_{\text{comm}}$ across nano-batches. The optimal $N$ depends on adapter composition, batch heterogeneity, accelerator characteristics, and network conditions.

Rather than relying on a static cost model, \name adapts the nano-batch size online using an
\emph{Additive-Increase/Multiplicative-Decrease (AIMD)} controller driven by end-to-end execution feedback.
Let $T_t$ denote the end-to-end batch completion time at scheduling horizon $t$.
Starting from a conservative nano-batch size $N_t$, \name updates
\begin{equation}
N_{t+1} =
\begin{cases}
N_t + \alpha, & \text{if } T_t \le T_{t-1} - \tau, \\
\max\{1, \lfloor \beta N_t \rfloor\}, & \text{otherwise},
\end{cases}
\end{equation}
where $\alpha$ is an additive step size ($\alpha{=}4$ by default), $\beta \in (0,1)$ is a multiplicative backoff factor ($\beta{=}1/2$ by default), and $\tau$ is a stability margin to filter noise.
Intuitively, \name increases granularity when finer pipelining reduces the critical path, and backs off when additional nano-batches elongate the step.
This feedback-driven adaptation converges quickly (in $O(\log N)$ steps to reduce from $N$ to 1). This implies negligible overhead relative to thousands of training iterations, especially noting that each adjustment step still makes forward training progress. 
Our evaluations show that our adaptive policy consistently outperforms manually tuned nano-batch sizes (\S\ref{subsec:ablation}).

\subsection{Adapter Scheduler: Maximizing Cluster Throughput}
\label{subsec:job-scheduler}

While the \emph{Model Fuser} and \emph{Kernel Fuser} optimize the training throughput for each LoRA group, the overall cluster throughput hinges on how multiple LoRA fine-tuning jobs are grouped and scheduled over time. Let each active job $j$ be characterized by a residual resource vector $r_j$ (e.g., unused GPU compute or memory) and an urgency score $u_j$ that captures its progress pressure, such as slowdown relative to standalone execution.
For any candidate group $G \subseteq \mathcal{J}$, we define their joint throughput $\widehat{T}(G)$.
The scheduler aims to identify groups that maximize $\widehat{T}(G)$ while ensuring that no job violates its progress constraint (e.g., completion deadlines or bounded slowdown), i.e.,
\begin{equation}
    \max_{G \subseteq \mathcal{J}} \; \widehat{T}(G)
\quad \text{s.t.} \quad
\forall j \in G,\; \Delta_j(G) \le \Delta_j^{\max},
\end{equation}
where $\Delta_j(G)$ denotes the slowdown incurred by job $j$ when executed in group $G$.
Directly solving this problem is intractable, since the space of possible groupings grows exponentially, and each grouping induces a different distributed execution plan with distinct performance characteristics. 


\name introduces a \emph{heterogeneity-aware Adapter Scheduler} that dynamically determines
(i) which LoRA jobs should be grouped,
(ii) when to merge or decouple existing groups, and
(iii) how to respect per-job progress requirements (e.g., completion deadlines or bounded slowdown).
Our key insight is that efficiency gains arise primarily from \emph{complementarity in residual resource usage}.
Jobs with unused compute or memory capacity can be paired with resource-hungry jobs to reduce idle time, whereas grouping similarly saturated jobs yields little benefit and often harms progress.
\name therefore focuses on exploiting such complementarities while explicitly avoiding harmful combinations.

\begin{algorithm}[t]
\caption{\name LoRA Batching Algorithm}
\label{alg:workflow}
\begin{algorithmic}[1]
\STATE \textbf{Input:} Job Queue $\mathcal{J}$, Cluster Resources $\mathcal{C}$

\WHILE{$\mathcal{J} \neq \emptyset$}
    \STATE \hspace{-.3cm} \textcolor{blue}{\footnotesize \itshape // \textbf{Phase 1: Adapter Scheduler} (\S\ref{subsec:job-scheduler}): Hierarchical Group.}
    \STATE \textcolor{gray}{\footnotesize \itshape // Sort jobs by Urgency (desc) and Residuals (asc)} \label{algo:group-start}
    \STATE $\mathcal{J} \leftarrow \textsc{Sort}(\mathcal{J} \mid u_j \downarrow, r_j \uparrow)$ \label{algo:sort}
    \STATE $j_{seed} \leftarrow \mathcal{J}.\text{pop\_front}()$
    
    \STATE \textcolor{gray}{\footnotesize \itshape // Find resource-complementary job to maximize group throughput}
    \STATE $k^* \leftarrow \operatorname*{arg\,max}_{k} \{\textsc{Throughput}( \{j_{seed}\} \cup \mathcal{J}[k] )\}$ \label{algo:find}
    
    \IF{$k^* > 0$}
        \STATE \textcolor{gray}{\footnotesize \itshape // Merge beneficial partners and re-insert for further grouping}
        \STATE $g_{new} \leftarrow \textsc{Merge}(\{j_{seed}\} \cup \mathcal{J}[k^*])$ \label{algo:merge}
        \STATE $\mathcal{J}.\text{insert}(g_{new})$; \quad $\mathcal{J} \leftarrow \mathcal{J} \setminus \mathcal{J}[k^*]$ \label{algo:reinsert}
        \STATE \textbf{continue} 
    \ELSE
        \STATE $g_{final} \leftarrow \{j_{seed}\}$ \label{algo:group-finish}
    \ENDIF

    \STATE \hspace{-.3cm} \textcolor{blue}{\footnotesize \itshape // \textbf{Phase 2: Model Fuser} (\S\ref{subsec:fuser}): SSM Compilation}
    \STATE $\mathcal{S}_{SSM} \leftarrow M_{base} \oplus \{ \text{Adapter}(j) \mid j \in g_{final} \}$ \label{algo:compile}
    \STATE $\Pi_{plan} \leftarrow \textsc{ParallelPlanner}(\mathcal{S}_{SSM}, \mathcal{C})$
    
    \STATE \hspace{-.3cm} \textcolor{blue}{\footnotesize \itshape // \textbf{Phase 3: Kernel Fuser} (\S\ref{subsec:kernel}): Adaptive Execution}
    \WHILE{step $t < T_{schedule}$}
        \STATE \textsc{FusedKernelLaunch}($\mathcal{S}_{SSM}, \Pi_{plan}, N_{nano}$) \label{algo:launch}
        \STATE $(\eta_{util}, \delta_{stall}) \leftarrow \textsc{Monitor\_Speed}()$
        \STATE \textsc{Kernel\_Update}($\eta_{util}$, $\delta_{stall}$) \label{algo:aimd}
    \ENDWHILE
    \STATE Update $\mathcal{J}$ with progress of $g_{final}$
\ENDWHILE
\end{algorithmic}
\end{algorithm}

Algorithm~\ref{alg:workflow} illustrates \name's batch LoRA training algorithm. For each LoRA job, we iteratively identify resource complementary partners that maximize joint throughput and group their execution (Line~\ref{algo:group-start}-\ref{algo:group-finish}). Once finalized, the group is compiled into a shared super-model, performing distributed execution (Line~\ref{algo:compile}). At the underlying per-iteration execution, \name exploits heterogeneity-aware LoRA fused kernels to adapt GPU block allocation to each LoRA for maximizing communication and computation overlap (Lines~\ref{algo:launch}--\ref{algo:aimd}). We next introduce how \name groups heterogeneous LoRAs into batch execution. 

\paragraph{Hierarchical Incremental Grouping.} 
To enable scalable grouping decisions, \name adopts a \emph{hierarchical incremental grouping policy} that progressively merges jobs with complementary utilization patterns (Line~\ref{algo:sort}).
Because grouping across broader resource tiers (e.g., across GPU nodes or ranks) incurs increasing communication overhead, \name follows a bottom-up strategy: it first forms groups within individual nodes, then considers merging across nodes, and finally across ranks.
Within each tier, grouping proceeds incrementally until additional merges no longer improve predicted efficiency; the resulting group is then finalized and lifted to the next tier.
This hierarchical process dramatically prunes the combinatorial search space, analogous to merge sort, while preserving the most beneficial grouping opportunities in practice.

When grouping within a tier, for each job (or an intermediate job group), \name maintains lightweight profiling statistics that capture residual hardware resources (\eg, GPU utilization) across its allocated machines. 
After identifying residual resources, \name sorts jobs by their residual resource availability in ascending order. 
Starting from the left-most (most resource-constrained) job, the scheduler performs a binary-cut search on the right-hand portion of the sorted list to find the cutoff where adding more jobs no longer improves the joint efficiency. The scheduler then forms the group, updates the grouped group's residual profile, and reinserts it into the queue. This incremental pack-and-reinsert loop repeats until no further beneficial merges are found, and each finalized group is compiled into an SSM (\S\ref{subsec:fuser}). This hierarchy dramatically prunes the combinatorial search space while preserving the most beneficial merges.

To prevent pathological slowdowns of individual jobs (i.e., ensuring $\Delta_j(G) \le \Delta_j^{\max}$), \name assigns each job an \emph{urgency score}, $\Delta_j(G)$, that reflects its proximity to violating a progress constraint, such as slowdown relative to standalone execution.
Jobs with higher urgency are given higher scheduling priority and placed earlier (leftmost) in the incremental grouping queue.
This ordering biases grouping decisions toward compensating progress-critical jobs with resource-abundant ones: jobs later in the queue expose greater residual capacity, which can be leveraged to accelerate earlier, more constrained jobs. 
In fact, our design enables \emph{elastic contribution}: jobs may temporarily release unused resources to benefit others and later grab more resources than their provisioned in isolation from other jobs to accelerate their own completion.
Our evaluation shows that this progress-aware scheduling improves not only aggregate training throughput but also per-job convergence time.

\paragraph{Complexity Analysis.}
For $K$ active jobs, exhaustive grouping is exponential in $K$.
In contrast, \name's hierarchical incremental policy runs in $O(K \log K)$ time per scheduling round: sorting costs $O(K \log K)$, and each merge involves $O(\log K)$ reinsertion.
In practice, this enables responsive scheduling in dynamic, online clusters while capturing the majority of attainable efficiency gains.

\section{Evaluation}
\label{sec:eval}

We implement \name on PyTorch 2.5 with Triton 3.5.1, developing a custom Triton kernel to efficiently support batched LoRA execution. \name natively supports existing distributed model execution (e.g., tensor parallelism) across GPUs without altering training semantics (i.e., lossless). Experimental results show that \name improves training throughput by 38\%, reduces per-job time-to-convergence by  $2.3\times$, and increases average GPU utilization by 37\%.

\subsection{Methodology}
\label{subsec:methodology}

\paragraph{Workloads, Models, and Tasks.}
Following prior advances~\cite{dlora-osdi24, mlora-vldb25}, we adopt a two-level evaluation methodology that combines micro-benchmarks on real hardware with large-scale trace-driven emulation. We run micro-benchmarks on a testbed equipped with 12 NVIDIA A100 GPUs, which offers accurate per-job LoRA training speed profiles too. Jobs arrive following the production GPU traces from ACMETrace~\cite{llmtrace-nsdi24}, which captures realistic multi-tenant cluster behavior, including job arrivals, GPU allocations, and execution durations, but does not natively include LoRA-specific attributes. To bridge this gap, we adopt LoRA configurations informed by prior LoRA serving studies~\cite{dlora-osdi24}. Concretely, for each job we randomly sample the LoRA rank from ${2,4,8,16}$ and the batch size from ${1,2,4,8}$ based on the original GPU allocation from the trace. Jobs randomly select either Llama-3-8B or Qwen-3-8B as base models. 

For large-scale evaluations, we use the collected per-job speed profile in the micro-benchmark, together with a production-grade distributed GPU training simulator~\cite{sailor-sosp25}. The simulator reports the training speed of different model architectures and hardware settings. We note that the simulator is widely used and the simulation error is within 3\% (Appendix~\ref{app:expset}). \name's optimization is lossless, does not alter the training accuracy. By default, jobs are trained on the GSM8K dataset, a math dataset of $\sim$8.5k grade-school-level questions. We report the evaluation results on a 128-GPU cluster by default. 

Due to the space limit, we include detailed experiment setup details (e.g., parameters) in Appendix~\ref{app:expset}. 

\paragraph{Baselines.}
We compare against the following advances:

\begin{denseitemize}
    \item \emph{mLoRA}~\cite{mlora-vldb25}: a state-of-the-art batched LoRA training system. mLoRA batches adapters using simple heuristics, such as grouping jobs as long as memory capacity permits, but does not explicitly account for heterogeneity across adapter jobs.

    \item \emph{Megatron}~\cite{megatron-sc21}: a widely used, highly optimized distributed LLM training framework. While Megatron provides efficient model parallelism, it trains each LoRA job independently.

    \item \emph{\name w/o Scheduler}: an ablation of \name that replaces the proposed Adapter Scheduler with mLoRA’s batching policy.

    \item \emph{\name w/o Kernel Fuser}: an ablation of \name that disables the fused heterogeneous LoRA kernel.
\end{denseitemize}

\paragraph{Metrics.}
We report three primary metrics: 
(i) \emph{Training Throughput} (samples/sec), defined as the cluster-wide throughput aggregated over all active jobs; 
(ii) \emph{Job Completion Time}, measured as the wall-clock time from job submission to training completion; and 
(iii) \emph{GPU Utilization}, computed as the average utilization across all GPUs.

All reported results are averaged over five independent runs.

\subsection{End-to-End Performance}
\label{subsec:e2e}


\begin{figure}[tp]
    \centering
    \begin{minipage}[t]{0.49\linewidth}
        \centering
        \includegraphics[width=\linewidth]{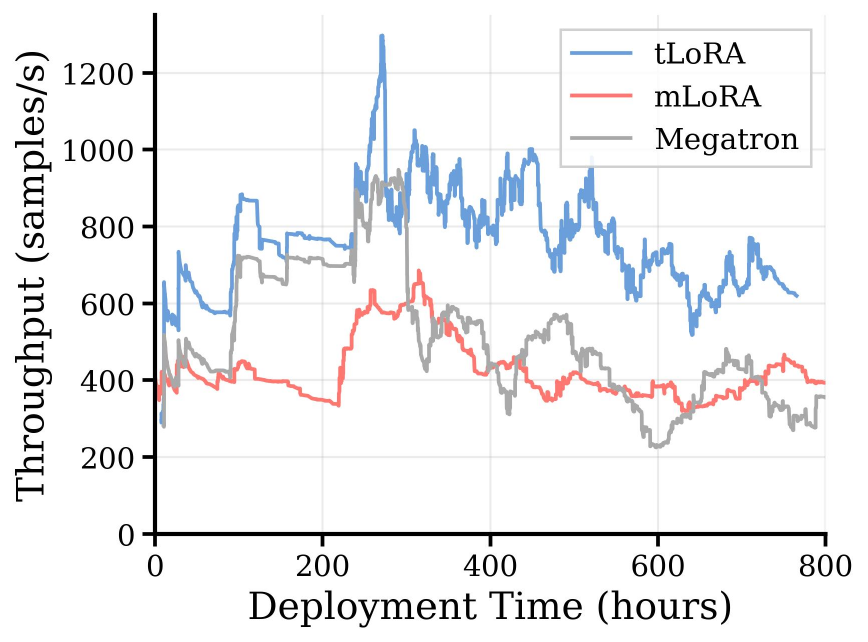}
        \vspace{-.5cm}
        \subcaption{Training throughput.}
        \label{fig:state-base-throughput}
    \end{minipage}\hfill
    \begin{minipage}[t]{0.49\linewidth}
        \centering
        \includegraphics[width=\linewidth]{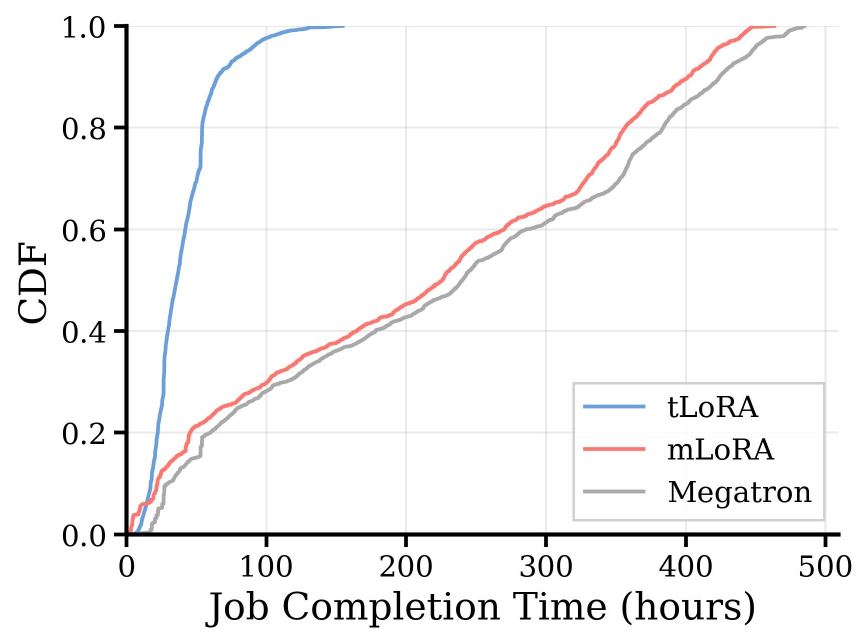}
       \vspace{-.5cm}
        \subcaption{Job completion time.}
        \label{fig:state-base-jct}
    \end{minipage}
    \caption{\name improves training throughput and completion.}
    \label{fig:state-base-exp}
\end{figure}

\paragraph{\name improves cluster training throughput.}
Figure~\ref{fig:state-base-throughput} shows that \name improves cluster-wide training throughput by 41\% over mLoRA under online workloads, where LoRA training jobs arrive dynamically and complete at different times. We observe two key effects. First, mLoRA often underperforms Megatron despite batching, as it groups jobs solely based on memory availability and ignores the communication overhead induced by co-location, which can negate compute gains. Second, \name consistently achieves higher throughput by combining SSM-based compilation with fused, rank-aware nano-batching, which jointly reduce pipeline bubbles and kernel launch overheads.


\paragraph{\name improves job training completion time.}
Figure~\ref{fig:state-base-jct} shows that \name reduces job completion time (JCT) by $5.4\times$ on average. Job completion time includes both queueing delay after submission and training convergence time during execution. These gains arise from two complementary mechanisms. First, \name co-locates jobs with complementary resource demands, enabling resource-abundant jobs to accelerate resource-constrained ones and improving cluster-wide utilization. Second, the progress-aware adapter scheduler dynamically prioritizes jobs experiencing slowdown under resource sharing, preventing starvation and eliminating long-tailed JCT. Ablation studies further confirm that \name consistently improves JCT across diverse real-world traces and system load conditions.

\begin{figure}[tp]
    \centering
    \begin{minipage}[t]{0.49\linewidth}
        \centering
        \includegraphics[width=\linewidth]{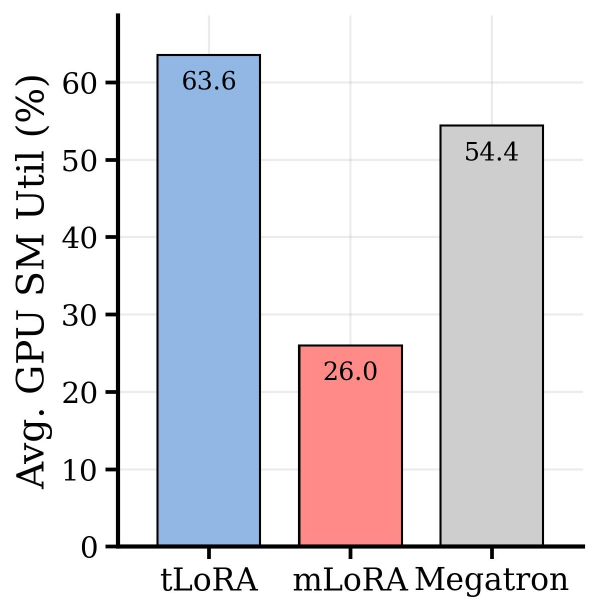}
        \vspace{-.5cm}
        \subcaption{Average GPU SM util.}
        \label{fig:gpu-util}
    \end{minipage}\hfill
    \begin{minipage}[t]{0.49\linewidth}
        \centering
        \includegraphics[width=\linewidth]{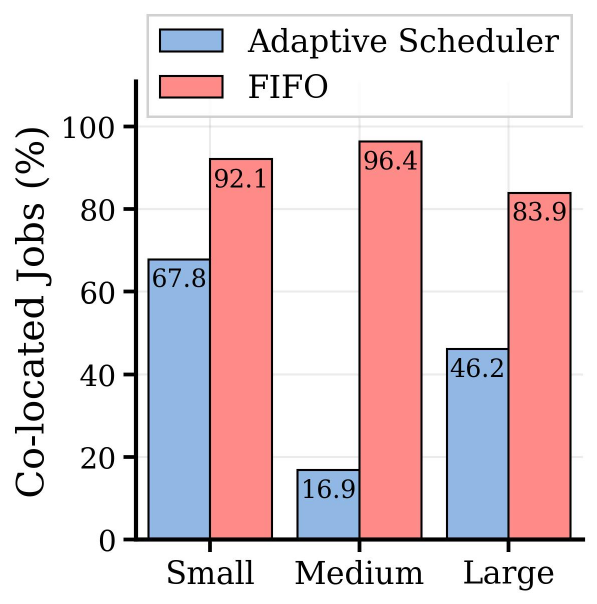}
        \vspace{-.5cm}
        \subcaption{Job colocation proportions.}
        \label{fig:coloc}
    \end{minipage}

    \caption{\name improves GPU utilization by effectively co-locating heterogeneous LoRA jobs.}
    \label{fig:gpu-coloc}
\end{figure}

\paragraph{\name improves GPU utilizations.}
Figure~\ref{fig:gpu-util} shows that \name improves GPU utilization up to 37\%, reducing idle/fragmented capacity. \name completes more training steps per GPU-hour by batching LoRA training jobs, thereby improving overall training throughput.

\begin{figure}[tp]
    \centering
    \begin{minipage}[t]{0.49\linewidth}
        \centering
        \includegraphics[width=\linewidth]{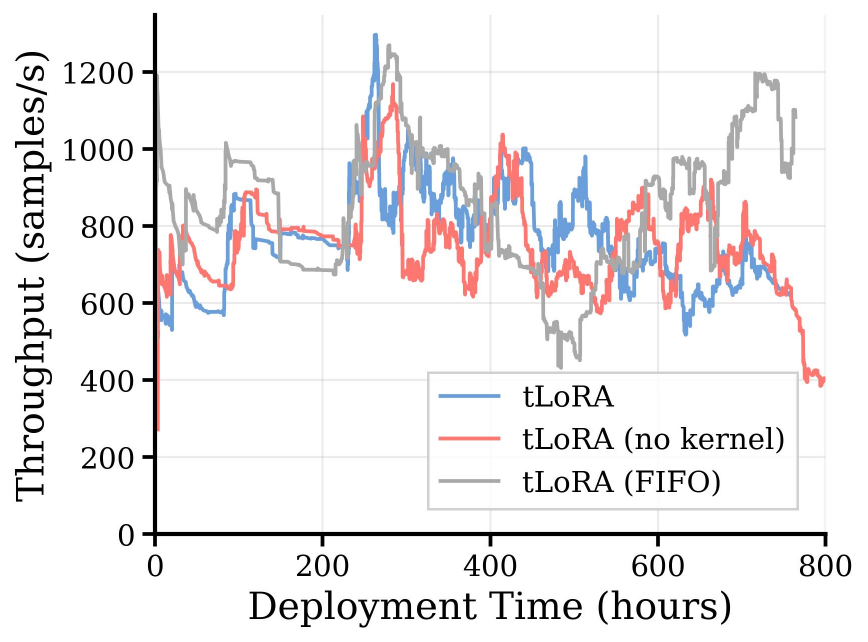}
        \vspace{-.5cm}
        \subcaption{Training throughput.}
        \label{fig:state-breakdown-throughput}
    \end{minipage}\hfill
    \begin{minipage}[t]{0.49\linewidth}
        \centering
        \includegraphics[width=\linewidth]{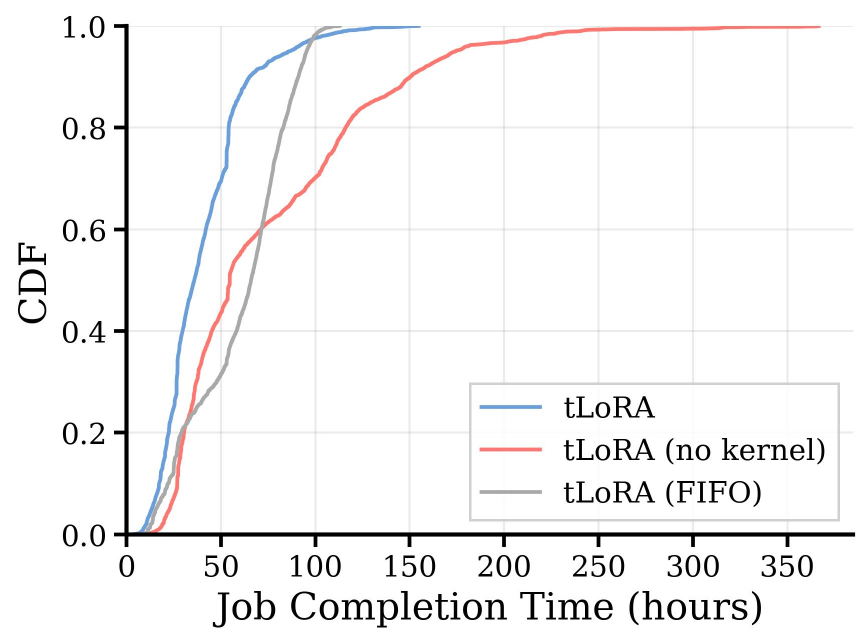}
        \vspace{-.5cm}
        \subcaption{Job completion time.}
        \label{fig:state-breakdown-jct}
    \end{minipage}

    \caption{Performance breakdown.}
    \label{fig:state-breakdown-exp}
\end{figure}


\subsection{Ablation Studies}
\label{subsec:ablation}

\paragraph{Performance Breakdown.} 
Figure~\ref{fig:coloc} dives into the grouping decisions across different jobs, where we analyze how jobs of different sizes are grouped with others. We define small jobs to be the jobs within bottom 33\% of compute cost based on their profiles (rank, batch size), medium to be the next third, and large to be the most costly 33\%. We notice that small and large jobs are mostly grouped, which is understandable as they are often complementary to each other. Similarly, jobs of medium sizes have a smaller grouping ratio, because they have limited idle resources and co-location benefit. In contrast, mLoRA's first-come-first-out (FIFO) policy naively co-locates jobs. Despite a higher grouping ratio, it achieves 5.4$\times$ slower training completion due to its much higher proportion of suboptimal co-location pairings. 


Beyond scheduling, Figure~\ref{fig:state-breakdown-exp} further isolates the contribution of \name's kernel-level optimizations. Replacing the fused heterogeneous LoRA kernel with the PyTorch-native kernel weakens the benefits of job co-location. In the unfused design, adapter updates repeatedly materialize small intermediate tensors and issue multiple per-adapter GEMMs, incurring high kernel launch overhead and poor data reuse. This fragmentation prevents effective overlap across adapters and amplifies execution bubbles.




\begin{figure}[t]
    \centering
    \begin{minipage}[t]{0.48\linewidth}
        \centering
        \includegraphics[width=\linewidth]{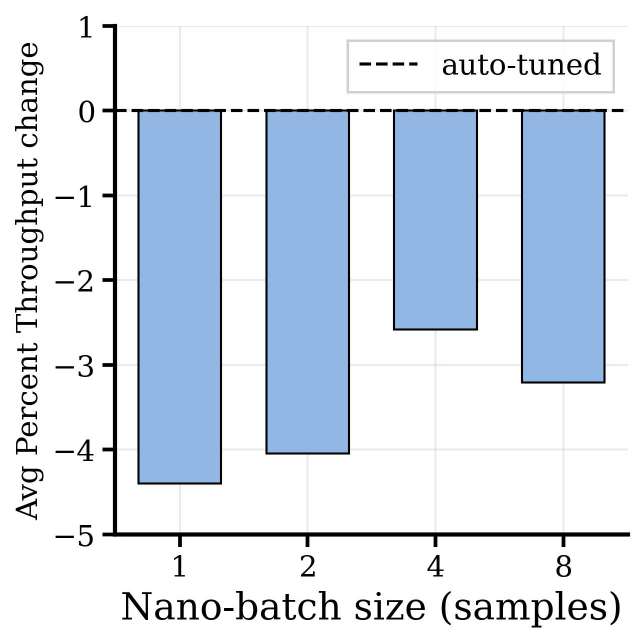}
        \vspace{-.5cm}
        \subcaption{Impact of nano-batch size.}
        \label{fig:ablation-nanoflow}
    \end{minipage}\hfill
    \begin{minipage}[t]{0.49\linewidth}
        \centering
        \includegraphics[width=\linewidth]{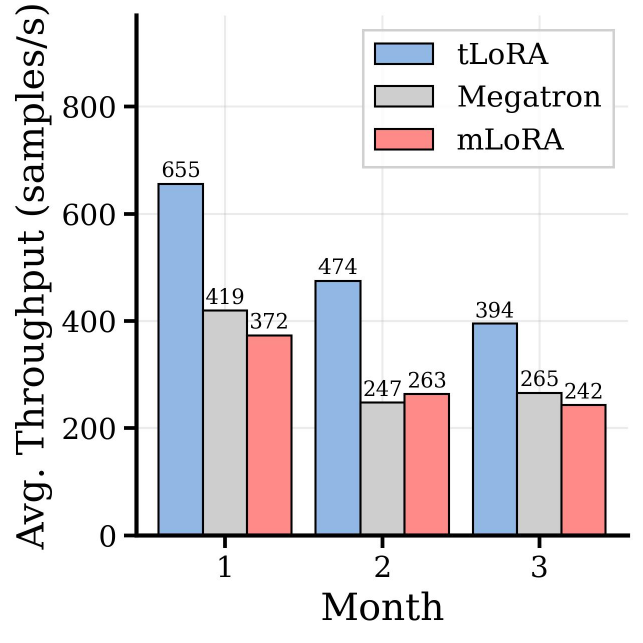}
        \vspace{-.5cm}
        \subcaption{Impact of job arrival pattern.}
        \label{fig:ablation-month}
    \end{minipage}
    \caption{Ablation studies on nano-batch size and arrival pattern. }
    \label{fig:ablation-nanoflow-month}
\end{figure}

\paragraph{Impact of nano-batch size.}
\name leverages an AIMD controller to decide the number of nano-batches, thereby optimizing GPU compute and communication overlap. Figure~\ref{fig:ablation-nanoflow} shows that compared to fixed nano-batch sizes, \name's adaptive design achieves higher end-to-end training throughput, confirming its effectiveness. 


\paragraph{Impact of online deployment traces.}
Figure~\ref{fig:ablation-month} evaluates \name under different arrival processes by replaying multiple months of ACMETrace while fixing job parameterization and cluster size. 
The first month exhibits the sparsest arrivals, resulting in shorter JCT since compatible co-location partners are more readily available; however, overall cluster throughput is slightly lower due to low contention and fewer opportunities for effective batching.
In contrast, Months~2 and~3 feature increasingly bursty arrivals, with approximately $2\times$ and $4\times$ higher job concurrency.
Despite this increased pressure, \name consistently sustains near-peak cluster throughput by adapting job grouping.


\begin{figure}[tp]
    \centering
    \begin{minipage}[t]{0.49\linewidth}
        \centering
        \includegraphics[width=\linewidth]{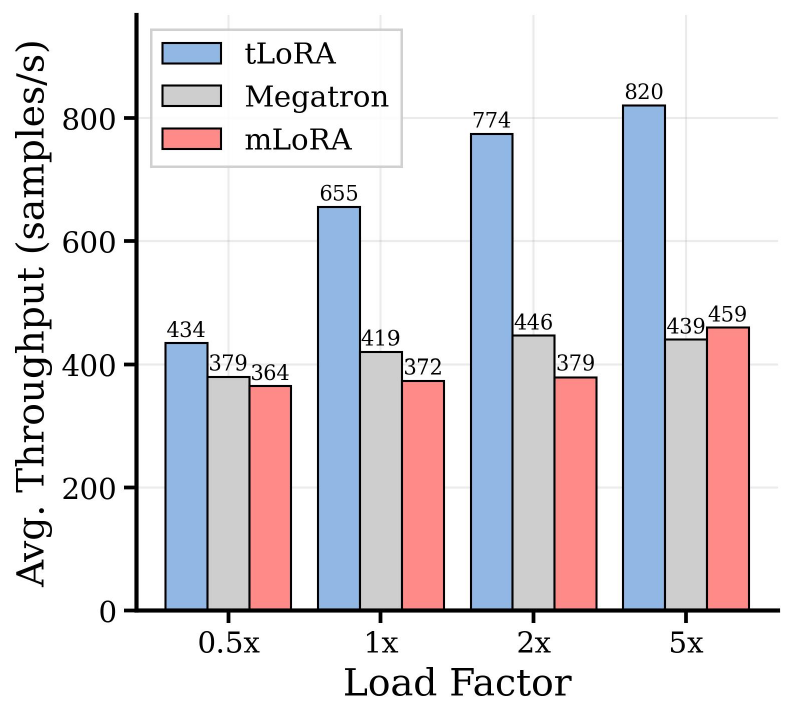}
        \vspace{-.5cm}
        \subcaption{Impact of scaling arrival rate.}
        \label{fig:ablation-scale}
    \end{minipage}\hfill
    \begin{minipage}[t]{0.49\linewidth}
        \centering
        \includegraphics[width=\linewidth]{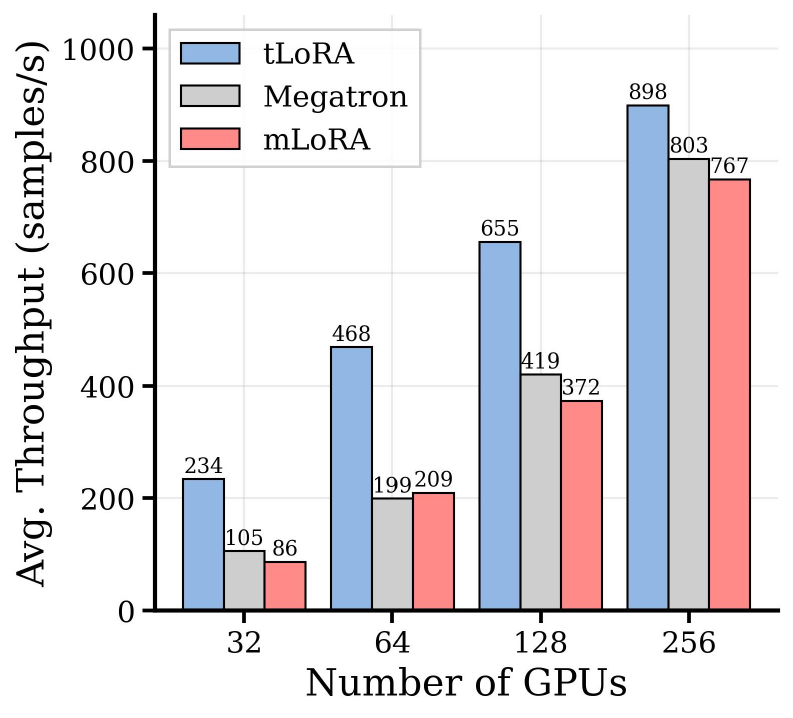}
        \vspace{-.5cm}
        \subcaption{Impact of cluster size.}
        \label{fig:ablation-cluster}
    \end{minipage}

    \caption{Ablation studies on arrival rate and cluster size.}
    \label{fig:ablation-scale-cluster}
\end{figure}

\paragraph{Impact of System Load.}
To stress-test the scheduler under varying load, we scale the job inter-arrival times by replaying the same trace with accelerated arrivals (e.g., $2\times$ and $5\times$ sooner) as well as slowed arrivals, as shown in Figure~\ref{fig:ablation-scale}. 
As arrivals become denser, queuing effects intensify and average job completion time increases.
Conversely, under sparser arrivals, cluster throughput decreases modestly due to underutilized batching opportunities, but job completion time improves. 
Importantly, \name achieves consistently 1.2--1.8$\times$ better throughput.

\paragraph{Impact of cluster size.}
Figure~\ref{fig:ablation-cluster} evaluates \name across clusters with varying numbers of available GPUs, using the same replayed workload and the simulator.
With fewer GPUs, \name maintains throughput proportional to the available capacity by continuously selecting runnable jobs and adapting grouping decisions to the tighter resource budget. As cluster size increases, \name scales to higher absolute throughput while preserving stable job completion times.
These results demonstrate that \name's performance benefits are robust across cluster provisioning levels.


\section{Related Work}
\label{sec:related}

\paragraph{LoRA Training and Serving Optimizations.}
Adapters are used to adapt large models by learning small additional modules rather than updating all weights. LoRA \cite{hu2021lora} introduced low-rank matrices to represent adaptation updates, significantly reducing memory usage and compute effort during fine-tuning. NanoFlow \cite{zhu2024nanoflow} focuses on serving systems; it splits inference requests into nano-batches and uses operation-level scheduling to overlap compute, memory, and network resources inside a device. Spindle \cite{wang2024spindle} addresses multi-task/multi-modal model training by sequencing tasks into execution waves and dynamically distributing computation to handle heterogeneity. In contrast, \name accounts for the heterogeneity of LoRA jobs to identify complementary job groups, while respecting per-job progress. 


\paragraph{Model Parallelism.}
Some works focus on improving parallelism performance, but they do not design fused LoRA adapter kernels. For example, Alpa \cite{zheng2022alpa} automates both inter- and intra-operator parallelism for model training by exploring search spaces of parallel execution and coordinating via a runtime system. NanoFlow \cite{zhu2024nanoflow} demonstrates intra-device overlap of operations inside a device for serving, but does not involve adapter training kernels or LoRA-type batching. Sailor \cite{sailor-sosp25} explores distributed training over dynamic and heterogeneous clusters, selecting device placements and training configurations via simulation and profiling, rather than focusing on GPU kernel design for adapter operations. These works contribute important insights into scheduling, placement, and resource use under heterogeneity, even though they do not include batched adapter kernels or interleaving of different adapter ranks inside a GPU in the way a LoRA-kernel-fusion work would.

\paragraph{ML Job Scheduling.}
Another group of work considers how to schedule and place training jobs in multi-GPU or multi-node settings. For example, Spindle \cite{wang2024spindle} uses wavefront scheduling to align computation waves across tasks, improving utilization under multi-task, multi-modal training workloads. Likewise, Sailor \cite{sailor-sosp25} uses configuration search and simulation to choose placements in heterogeneous clusters, optimizing throughput and cost under dynamic hardware and network conditions. These contributions show how scheduling and placement choices affect efficiency across jobs, though they generally assume that execution kernels or adapter behaviors are given rather than co-optimized with scheduling.



\section{Conclusion}

This paper presents \name, a heterogeneity-aware framework for efficiently training multiple LoRA adapters in shared GPU clusters. By unifying concurrent tuning jobs through a Shared Super-Model abstraction, introducing a fused heterogeneous LoRA kernel with adaptive nano-batching, and coordinating jobs via a progress-aware scheduler, \name transforms multi-LoRA training from isolated jobs into a jointly optimized learning workload. Our design preserves the semantics and convergence behavior of independent training while substantially improving collective throughput, GPU utilization, and per-job completion time under realistic multi-tenant traces. 

\bibliographystyle{icml2026}
\bibliography{refs.bib}

\begin{thebibliography}{14}
\providecommand{\natexlab}[1]{#1}
\providecommand{\url}[1]{\texttt{#1}}
\expandafter\ifx\csname urlstyle\endcsname\relax
  \providecommand{\doi}[1]{doi: #1}\else
  \providecommand{\doi}{doi: \begingroup \urlstyle{rm}\Url}\fi

\bibitem[Hu et~al.(2022)Hu, Shen, Wallis, Allen-Zhu, Li, Wang, and Chen]{hu2021lora}
Hu, E.~J., Shen, Y., Wallis, P., Allen-Zhu, Z., Li, Y., Wang, L., and Chen, W.
\newblock {LoRA}: Low-rank adaptation of large language models.
\newblock In \emph{ICLR}, 2022.
\newblock \url{https://arxiv.org/abs/2106.09685}.

\bibitem[Hu et~al.(2024)Hu, Ye, Wang, Wang, Zhang, Chen, Sun, Lin, Wang, Luo, Wen, and Zhang]{llmtrace-nsdi24}
Hu, Q., Ye, Z., Wang, Z., Wang, G., Zhang, M., Chen, Q., Sun, P., Lin, D., Wang, X., Luo, Y., Wen, Y., and Zhang, T.
\newblock Characterization of large language model development in the datacenter.
\newblock In \emph{NSDI}, 2024.
\newblock \url{https://arxiv.org/abs/2403.07648}.

\bibitem[Luo et~al.(2024)Luo, Wong, Trabucco, Huang, Gonzalez, Chen, Salakhutdinov, and Stoica]{stylus-arxiv24}
Luo, M., Wong, J., Trabucco, B., Huang, Y., Gonzalez, J.~E., Chen, Z., Salakhutdinov, R., and Stoica, I.
\newblock {Stylus}: Automatic adapter selection for diffusion models.
\newblock In \emph{NeurIPS}, 2024.
\newblock \url{https://arxiv.org/abs/2404.18928}.

\bibitem[Narayanan et~al.(2021)Narayanan, Shoeybi, Casper, LeGresley, Patwary, Korthikanti, Vainbrand, Kashinkunti, Bernauer, Catanzaro, Phanishayee, and Zaharia]{megatron-sc21}
Narayanan, D., Shoeybi, M., Casper, J., LeGresley, P., Patwary, M., Korthikanti, V., Vainbrand, D., Kashinkunti, P., Bernauer, J., Catanzaro, B., Phanishayee, A., and Zaharia, M.
\newblock Efficient large-scale language model training on gpu clusters using megatron-lm.
\newblock In \emph{SC}, 2021.
\newblock \url{https://arxiv.org/abs/2104.04473}.

\bibitem[Sheng et~al.(2024)Sheng, Cao, Li, Hooper, Lee, Yang, Chou, Zhu, Zheng, Keutzer, Gonzalez, and Stoica]{slora-mlsys24}
Sheng, Y., Cao, S., Li, D., Hooper, C., Lee, N., Yang, S., Chou, C., Zhu, B., Zheng, L., Keutzer, K., Gonzalez, J.~E., and Stoica, I.
\newblock {S-LoRA}: Serving thousands of concurrent {LoRA} adapters.
\newblock In \emph{MLSys}, 2024.
\newblock \url{https://arxiv.org/abs/2311.03285}.

\bibitem[Strati et~al.(2025)Strati, Zhang, Manos, {S\'anchez P\'eriz}, Hu, Chen, Buzcu, Han, Delgado, and Klimovic]{sailor-sosp25}
Strati, F., Zhang, Z., Manos, G., {S\'anchez P\'eriz}, I., Hu, Q., Chen, T., Buzcu, B., Han, S., Delgado, P., and Klimovic, A.
\newblock {Sailor}: Automating distributed training over dynamic, heterogeneous, and geo-distributed clusters.
\newblock In \emph{SOSP}, 2025.
\newblock \url{https://arxiv.org/abs/2504.17096}.

\bibitem[Um et~al.(2024)Um, Oh, Kang, Lee, Kim, Kim, Kim, Muzzammil, and Jeon]{metis-atc24}
Um, T., Oh, B., Kang, M., Lee, W.-Y., Kim, G., Kim, D., Kim, Y., Muzzammil, M., and Jeon, M.
\newblock Metis: Fast automatic distributed training on heterogeneous gpus.
\newblock In \emph{ATC24}, 2024.
\newblock \url {https://www.usenix.org/system/files/atc24-um.pdf}.

\bibitem[Wang et~al.(2025)Wang, Zhu, Fu, Miao, Zhang, Zhu, Hong, Li, and Cui]{wang2024spindle}
Wang, Y., Zhu, S., Fu, F., Miao, X., Zhang, J., Zhu, J., Hong, F., Li, Y., and Cui, B.
\newblock {Spindle}: Efficient distributed training of multi-task large models via wavefront scheduling.
\newblock In \emph{ASPLOS}, 2025.
\newblock \url{https://arxiv.org/abs/2409.03365}.

\bibitem[Wu et~al.(2024)Wu, Zhu, Zhang, Sun, Liu, and Jin]{dlora-osdi24}
Wu, B., Zhu, R., Zhang, Z., Sun, P., Liu, X., and Jin, X.
\newblock {dLoRA}: Dynamically orchestrating requests and adapters for {LoRA} {LLM} serving.
\newblock In \emph{OSDI}, 2024.
\newblock \url{https://www.usenix.org/system/files/osdi24-wu-bingyang.pdf}.

\bibitem[Wu et~al.(2025)Wu, Piao, Huang, Wang, Li, Pfister, Meng, Ma, and Wei]{sdlora-iclr2025}
Wu, Y., Piao, H., Huang, L.-K., Wang, R., Li, W., Pfister, H., Meng, D., Ma, K., and Wei, Y.
\newblock Sd-lora: Scalable decoupled low-rank adaptation for class incremental learning.
\newblock In \emph{ICLR}, 2025.
\newblock \url{https://arxiv.org/abs/2501.13198}.

\bibitem[Ye et~al.(2025)Ye, Li, Hu, Lan, Sha, Zhang, Duan, Zuo, Lu, Zhou, and Tang]{mlora-vldb25}
Ye, Z., Li, D., Hu, Z., Lan, T., Sha, J., Zhang, S., Duan, L., Zuo, J., Lu, H., Zhou, Y., and Tang, M.
\newblock mlora: Fine-tuning lora adapters via highly-efficient pipeline parallelism in multiple gpus.
\newblock 2025.
\newblock \url{https://arxiv.org/abs/2312.02515}.

\bibitem[Zhang et~al.(2023)Zhang, Chen, Bukharin, Karampatziakis, He, Cheng, Chen, and Zhao]{adalora-iclr2023}
Zhang, Q., Chen, M., Bukharin, A., Karampatziakis, N., He, P., Cheng, Y., Chen, W., and Zhao, T.
\newblock Adalora: Adaptive budget allocation for parameter-efficient fine-tuning.
\newblock In \emph{ICLR}, 2023.
\newblock \url{https://arxiv.org/abs/2303.10512}.

\bibitem[Zheng et~al.(2022)Zheng, Li, Zhang, Zhuang, Chen, Huang, Wang, Xu, Zhuo, Xing, Gonzalez, and Stoica]{zheng2022alpa}
Zheng, L., Li, Z., Zhang, H., Zhuang, Y., Chen, Z., Huang, Y., Wang, Y., Xu, Y., Zhuo, D., Xing, E.~P., Gonzalez, J.~E., and Stoica, I.
\newblock {Alpa}: Automating inter- and intra-operator parallelism for distributed deep learning.
\newblock In \emph{OSDI}, 2022.
\newblock \url{https://arxiv.org/abs/2201.12023}.

\bibitem[Zhu et~al.(2025)Zhu, Zhao, Zhao, Zuo, Gu, Xie, Gao, Xu, Tang, Ye, Kamahori, Lin, Wang, Krishnamurthy, and Kasikci]{zhu2024nanoflow}
Zhu, K., Zhao, Y., Zhao, L., Zuo, G., Gu, Y., Xie, D., Gao, Y., Xu, Q., Tang, T., Ye, Z., Kamahori, K., Lin, C.-Y., Wang, S., Krishnamurthy, A., and Kasikci, B.
\newblock {NanoFlow}: Towards optimal large language model serving throughput.
\newblock In \emph{OSDI}, 2025.
\newblock \url{https://arxiv.org/abs/2408.12757}.

\end{thebibliography}

\clearpage
\appendix
\begin{figure}[tp]
    \centering
        \centering
        \includegraphics[width=0.50\linewidth]{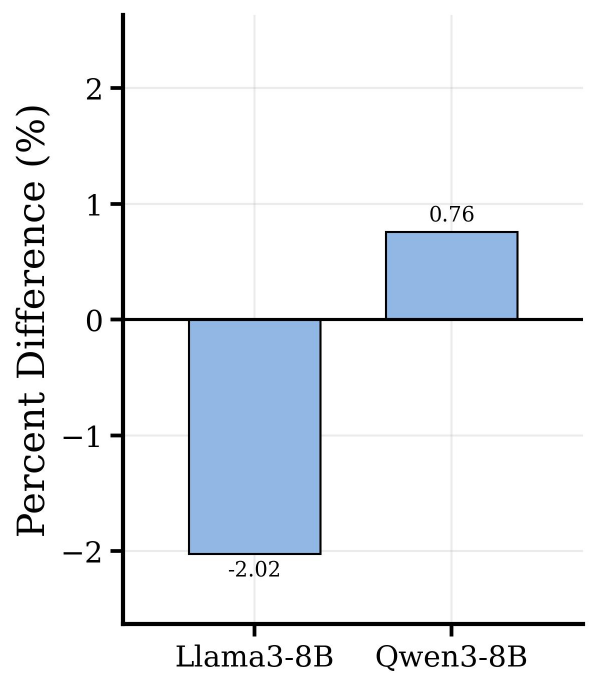}
        \vspace{-4pt}
        \caption{Simulator average iteration accuracy by model.}
        \label{fig:sim}
\end{figure}

\begin{figure}[tp]
    \centering
        \centering
        \includegraphics[width=0.50\linewidth]{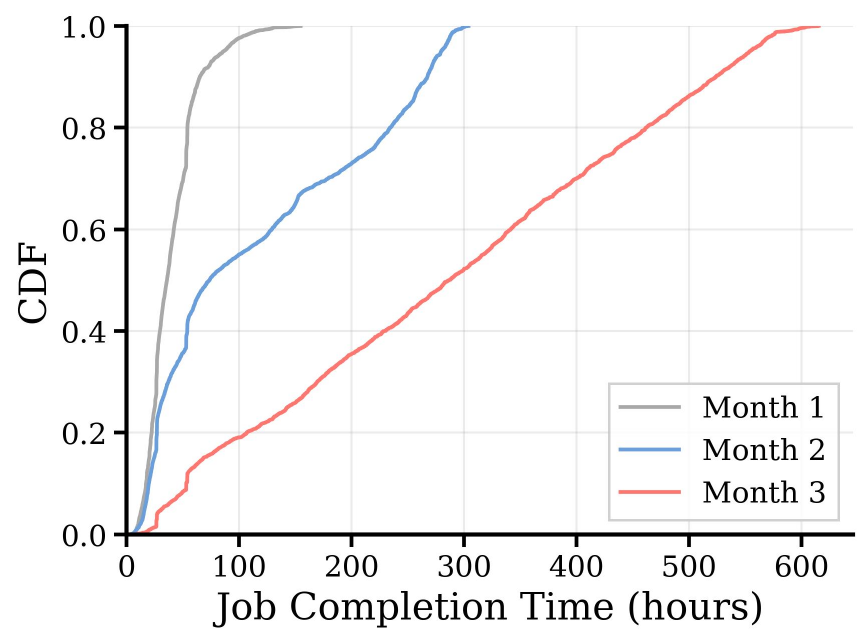}
        \vspace{-4pt}
        \caption{Impact of arrival trace on job completion time.}
        \label{fig:jct-month}
\end{figure}

\begin{figure}[tp]
    \centering
        \centering
        \includegraphics[width=0.50\linewidth]{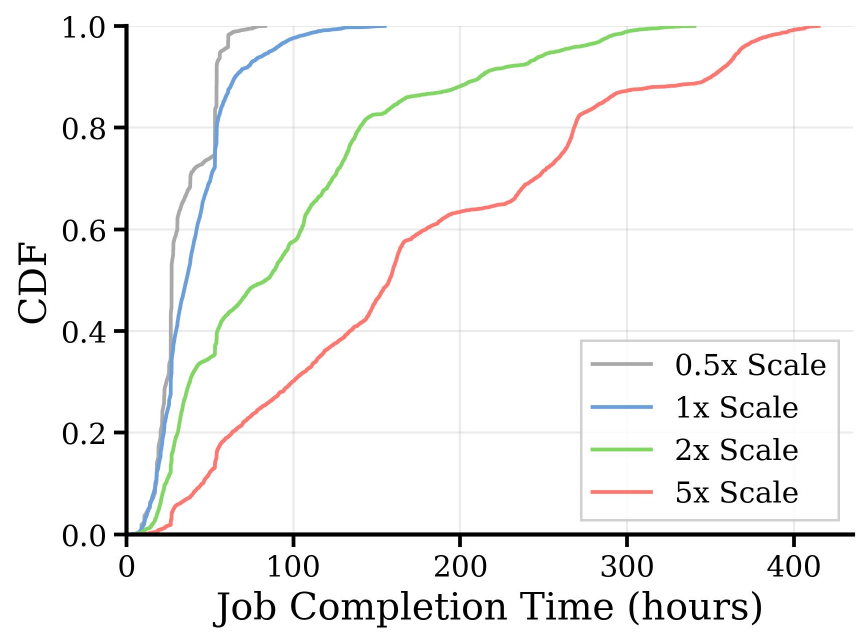}
        \vspace{-4pt}
        \caption{Impact of scaling arrival rate on job completion time.}
        \label{fig:jct-scale}
\end{figure}

\begin{figure}[tp]
    \centering
        \centering
        \includegraphics[width=0.50\linewidth]{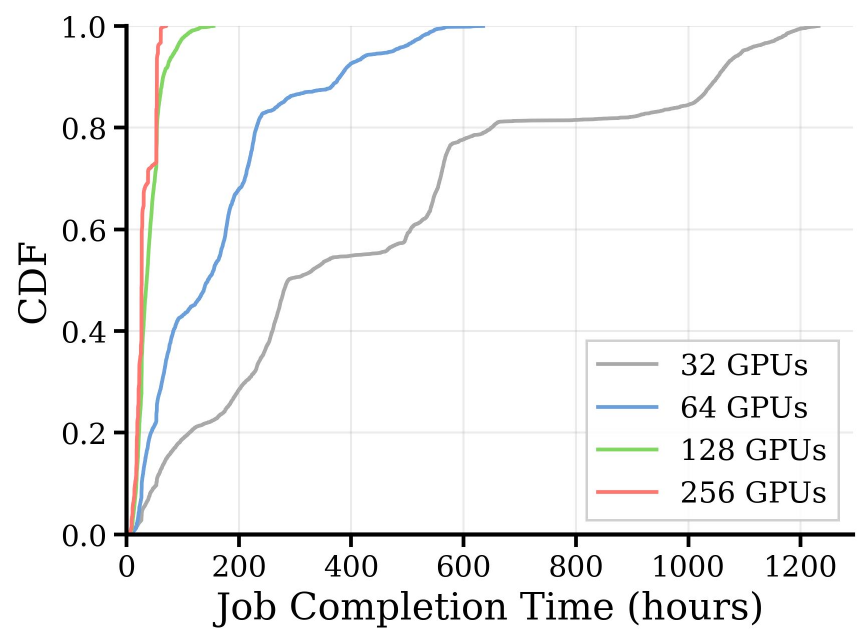}
        \vspace{-4pt}
        \caption{Impact of cluster size on job completion time.}
        \label{fig:jct-cluster}
\end{figure}

\section{Appendix}

\subsection{Experiment settings}
\label{app:expset}

Each adapter specification (rank $r$, batch size, maximum sequence length, step budget, and GPU count) is fixed at submission time and not altered by \name. By default, we run on a cluster size of 128 GPUs using the first month of data from the trace\_seren.csv provided by ACMETrace. 
We use micro-benchmarks from the Sailor training simulator to extrapolate training times and find beneficial co-location pairs. Because the simulator runs real forward and backwards passes on layers of the model using Megatron-LM for model parallelism, then extrapolates execution time of the whole model by taking advantage of the homogeneity of transformer layers, we find that it is very accurate, as shown in Figure~\ref{fig:sim}.

The cluster enforces a global concurrency cap of up to 128 runnable jobs based on GPU availability.
When several jobs arrive simultaneously, \name must (i) decide which jobs start immediately, (ii) decide whether not to co-locate adapters with other compatible adapters available.
Compatible adapters are those whose memory footprints and sequence lengths can be jointly satisfied without violating device memory limits.
We measure per-iteration time once job group placement is stable and the GPU is warmed up.

\subsection{Job completion time ablation studies}
\label{app:ablations}

Similar to \S\ref{subsec:ablation}, we show that \name maintains its performance under varying load conditions. 

\paragraph{Impact of online deployment traces}
Months 2 and 3 contain much denser bursts of jobs when compared to Month 1, leading to some reduction in average job completion time as jobs must queue for available GPU resources, even with co-location. We see that the cluster GPU resources are quickly saturated by much denser job arrival patterns leading to a flatter job completion time shown in Figure~\ref{fig:jct-month}. However, as noted in \S\ref{subsec:ablation}, \name maintains near-peak throughput, indicating that it is still able to make optimal co-locating and scheduling decisions despite the much longer job queues. 

\paragraph{Impact of System Load.}
Similarly, when scaling the job arrival rate, high arrival rates lead to longer job completion times as the available computing resources are more quickly saturated. This is reflected in the much flatter job completion time curves shown in Figure~\ref{fig:jct-scale} for 2$\times$ and 5$\times$ arrival rates. On the other hand, when arrivals are slowed, the average job completion time does not suffer, as seen by the job completion CDF of the 0.5$\times$ curve being slightly better than the default. 

\paragraph{Impact of cluster size.}
Finally, even with reduced cluster sizes, \name is able to maintain proportional job completion time, as it more intelligently selects jobs to co-locate and execute to maximize efficiency with limited resources. Figure~\ref{fig:jct-cluster} confirms this scalability; as the cluster size decreases from 256 to 32 GPUs, the completion time curves shift rightward in consistent, predictable intervals rather than diverging exponentially. While the slope naturally flattens for the 32-GPU configuration due to lower resource availability, the steady gradient across all scenarios indicates that the scheduler successfully avoids the resource starvation and heavy-tail latency often seen in constrained environments.

\end{document}